\PassOptionsToPackage{table}{xcolor}
\documentclass{hustmedia-paper}
\ifPDFTeX
  \usepackage[utf8]{inputenc}
\fi
\setcopyright{none}
\HMMAuthorLine{ZeLiang Chen\textsuperscript{*}, Yunyao Zhang\textsuperscript{*}, Xinglang Zhang, Wenbing Li, Junqing Yu, Zikai Song\textsuperscript{\ensuremath{\dagger}}}
\HMMAffiliationLine{Huazhong University of Science and Technology}
\HMMAuthorNote{\textsuperscript{*}Equal contribution\qquad\textsuperscript{\ensuremath{\dagger}}Corresponding author}

\usepackage{amsmath}
\usepackage{graphicx}
\usepackage{hyperref}
\let\mathsfit\relax
\ifPDFTeX
  
\fi
\usepackage{amsmath,amsfonts,bm}

\def\eqref#1{equation~\ref{#1}}
\def\1{\bm{1}}

\DeclareMathAlphabet{\mathsfit}{\encodingdefault}{\sfdefault}{m}{sl}
\SetMathAlphabet{\mathsfit}{bold}{\encodingdefault}{\sfdefault}{bx}{n}

\usepackage{wrapfig}
\usepackage{url}
\usepackage{xcolor}
\usepackage{booktabs}
\usepackage{multirow}
\usepackage{makecell}
\usepackage{enumitem}
\ifPDFTeX
  
\fi
\usepackage{amssymb}
\usepackage{subcaption}
\usepackage{colortbl}
\usepackage{algorithm}
\usepackage{algorithmicx}
\usepackage{algpseudocode}
\usepackage{amsthm}
\usepackage{fvextra}
\ifPDFTeX
  \DeclareUnicodeCharacter{2227}{\ensuremath{\land}}
  \DeclareUnicodeCharacter{2228}{\ensuremath{\lor}}
  \DeclareUnicodeCharacter{00AC}{\ensuremath{\neg}}
  \DeclareUnicodeCharacter{27E8}{\ensuremath{\langle}}
  \DeclareUnicodeCharacter{27E9}{\ensuremath{\rangle}}
\else
\begingroup
\catcode`∧=\active
\catcode`∨=\active
\catcode`⟨=\active
\catcode`⟩=\active
\gdef\HMMVerbatimSymbols{%
  \def∧{\makebox[\fontcharwd\font`x][c]{\ensuremath{\land}}}%
  \def∨{\makebox[\fontcharwd\font`x][c]{\ensuremath{\lor}}}%
  \def⟨{\makebox[\fontcharwd\font`x][c]{\ensuremath{\langle}}}%
  \def⟩{\makebox[\fontcharwd\font`x][c]{\ensuremath{\rangle}}}%
}
\endgroup
\fvset{codes*={\catcode`∧=\active\catcode`∨=\active
  \catcode`⟨=\active\catcode`⟩=\active},defineactive*=\HMMVerbatimSymbols}
\fi
\tcbuselibrary{most}

\theoremstyle{definition}

\theoremstyle{plain}

\newcommand{\best}[1]{\textbf{#1}}
\newcommand{\second}[1]{\underline{#1}}
\newcommand{\posdelta}[1]{\textcolor{green!50!black}{\scriptsize #1}}
\newcommand{\negdelta}[1]{\textcolor{red!80!black}{\scriptsize #1}}
\definecolor{CMTgreen}{RGB}{35,135,75}
\newcommand{\cmt}[1]{\textcolor{CMTgreen}{\quad $\triangleright$ #1}}

\newcommand{\qwen}{\raisebox{-0.16em}{\includegraphics[height=1.15em]{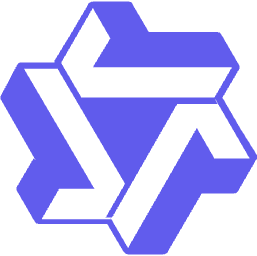}}}
\newcommand{\llama}{\raisebox{-0.16em}{\includegraphics[height=1.15em]{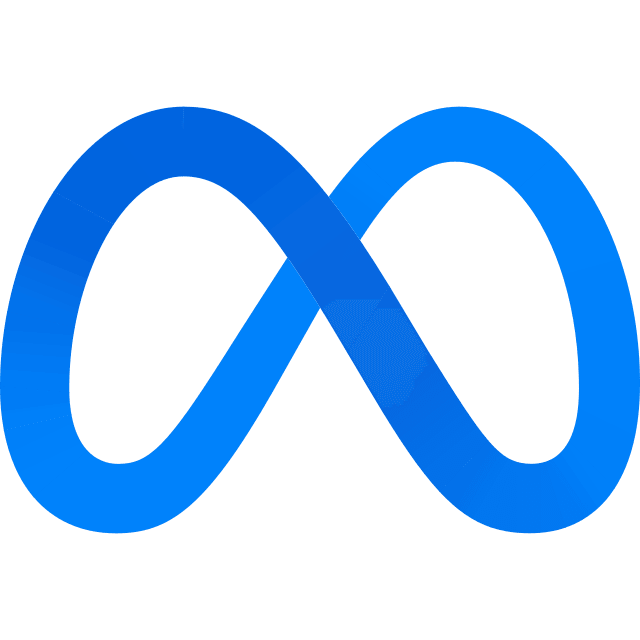}}}
\usepackage[nameinlink]{cleveref}

\title{VeriLogic: Reinforcement Learning over Verifiable Logical Reasoning Traces}
\HMMProjectWebsite{https://ai4ss.github.io/VeriLogic/}
\HMMPublicationStatus{Preprint}
\HMMCorrespondingEmail{skyesong@hust.edu.cn}
\author{ZeLiang Chen}
\author{Yunyao Zhang}
\author{Xinglang Zhang}
\author{Wenbing Li}
\author{Junqing Yu}
\author{Zikai Song}
\begin{abstract}
Logical reasoning derives conclusions from premises through explicit intermediate inference steps, providing a natural basis for process-level verification and supervision. 
However, existing learning methods often rely on response-level outcome signals, which overlook heterogeneous correctness within the reasoning trace: a correct answer may contain invalid reasoning, while valid intermediate steps may also appear in an incorrect response.
To address this issue, we introduce Reinforcement Learning on \textbf{Veri}fiable \textbf{Logic}al Reasoning (\textbf{VeriLogic}), which turns internal reasoning traces into explicit and verifiable learning objects. 
Rather than treating a response as an indivisible learning unit, VeriLogic decomposes logical reasoning into explicit intermediate judgments, proofs, and final decisions, and organizes them into a unified ``\textbf{Grounding - Optimization - Refinement}'' pipeline: 
(1) \textit{\textbf{Verification-Grounded Reasoning Environment Construction}} grounds logical reasoning into explicit intermediate states and formally checkable proof obligations, making the internal reasoning process accessible to verification and learning; 
(2) \textit{\textbf{Logical Structured Policy Optimization}} learns from verifier feedback through complementary global and local rewards, jointly optimizing the overall reasoning outcome while providing targeted optimization signals to different intermediate reasoning paths; and 
(3) \textit{\textbf{Verifier-Guided Iterative Refinement}} acts on verifier feedback by leveraging proof states and error messages to iteratively diagnose and repair failed reasoning.
Experiments on six logical reasoning benchmarks show consistent gains over strong baselines, with average improvements of 6.39\% and 5.00\% on Qwen3 and Llama-3.1. These results highlight the value of making logical reasoning traces verifiable, learnable, and refinable beyond final-answer supervision.
\end{abstract}

\begin{document}
\maketitle

\section{Introduction}
% Reliable logical reasoning requires correct judgments supported by valid evidence. Given a set of premises, a large language model (LLM) must determine whether a proposition is supported, refuted, or left undetermined, and justify any claim of derivability. Chain-of-thought prompting and reinforcement learning have substantially improved LLM reasoning performance \citep{wei2022chain,guo2025deepseekr1}. Yet final-answer correctness does not establish the validity of the reasoning behind it. As illustrated in Figure~\ref{fig:motivation}, a model can reach the correct judgment while failing to provide a proof that passes formal verification. Training for reliable logical reasoning must therefore account for both judgment correctness and proof validity.

% background
Logical reasoning~\citep{smith2003introduction,enderton2001mathematical} is a fundamental capability of LLMs~\citep{huang2023towards,liu2025logical}, where a model derives conclusions from given premises through valid inference. 
Unlike many open-ended reasoning tasks~\citep{cobbe2021training,hendrycks2021measuring}, logical reasoning is \textit{\textbf{naturally structured}}: intermediate judgments can be decomposed, inference paths can be traced, and their validity can often be explicitly checked. Yet this structure \textit{\textbf{remains largely underused}}. 
Existing methods mainly improve logical reasoning through prompting~\citep{wei2022chain,zhou2022least} or inference-time augmentation~\citep{wang2022self,yao2023tree}, rather than internalizing it as a trainable model capability.
Reinforcement Learning on Logic Reasoning offers a promising alternative~\citep{wang2025a2rl}, but a few current attempts~\citep{xie2025logic} predominantly optimize final outputs with response-level rewards, overlooking the heterogeneous correctness that may arise across intermediate reasoning paths~\citep{wei2025curabl}.

\begin{figure*}[t!]
\centering
\includegraphics[width=0.92\textwidth]{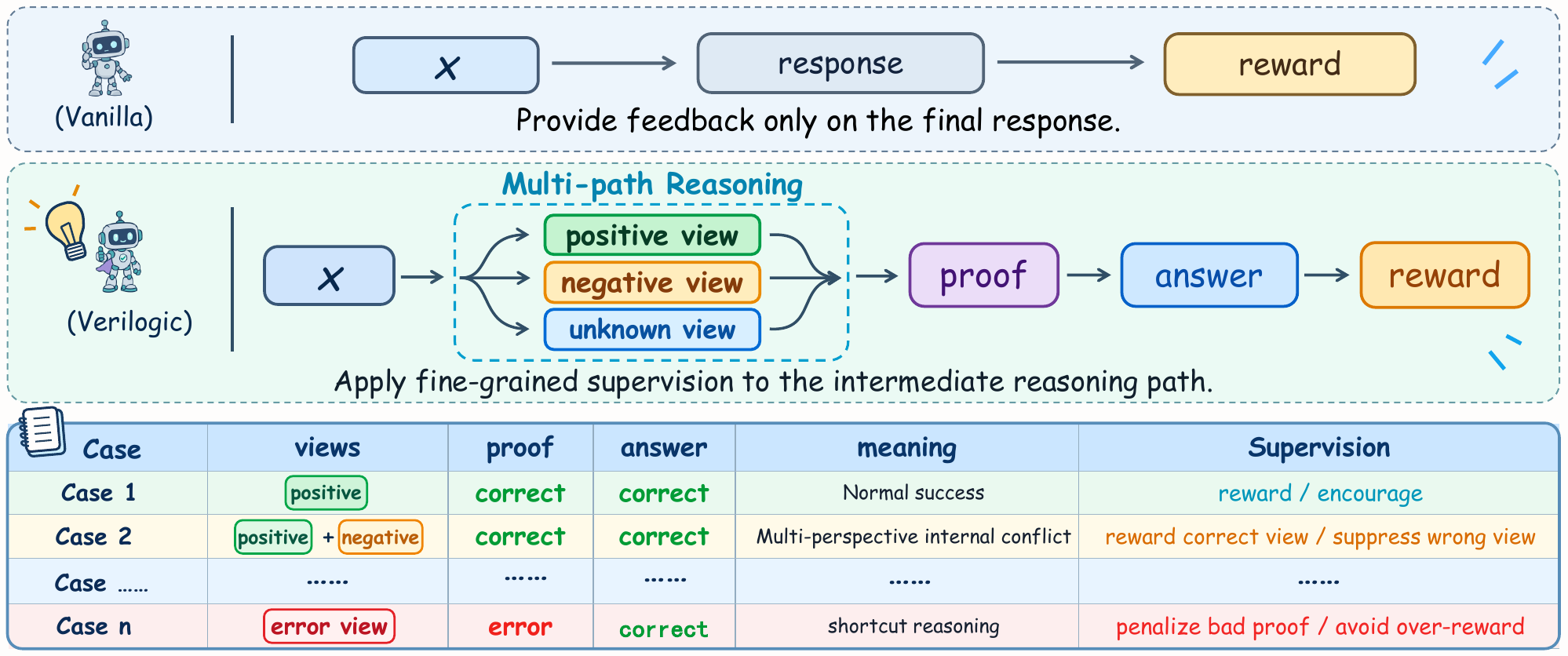}
\caption{\textbf{Heterogeneous correctness in logical reasoning.}
A final answer may conceal errors in intermediate reasoning views or proofs, while fine-grained supervision distinguishes and corrects these cases.}
\label{fig:motivation}
\end{figure*}

% motivation
We exploit the potential capability of logical reasoning is to make its internal process both \textbf{traceable} and \textbf{verifiable}.
Instead of treating reasoning as a single response, we break it into multiple explicit reasoning traces. Each trace follows a distinct reasoning path with intermediate judgments that can be examined step by step.
More importantly, these traces can be optimized with verifier feedback according to their own correctness, rather than sharing a single reward from the final answer. This lets the model distinguish different reasoning outcomes—for example, correct reasoning with a wrong answer, wrong reasoning with a correct answer, or cases where both succeed or fail. As shown in Figure \ref{fig:motivation}.
Motivated by this view, we introduce Reinforcement Learning on \textbf{Veri}fiable \textbf{Logic}al Reasoning (\textbf{VeriLogic}), which turns internal reasoning traces into explicit and verifiable learning objects for fine-grained policy optimization.

% challenge
Turning traceable and verifiable reasoning into an effective RL framework requires addressing three challenges: 
(\textbf{C1}) \textbf{\textit{How to construct a verifiable reasoning space?}} Different reasoning paths must be explicitly represented, semantically aligned with the original problem, and formally checkable;
(\textbf{C2}) \textit{\textbf{How to optimize reasoning at the right granularity?}} Different reasoning paths within the same response may have different correctness states, so a single response-level reward cannot provide sufficiently precise optimization signals;
(\textbf{C3}) \textbf{\textit{How to make verification actionable after failure?}} Even after training, a model may still produce invalid or incomplete reasoning, requiring verifier feedback to guide iterative revision rather than serving only as a final correctness check.

To address these challenges, VeriLogic organizes logical reasoning into a unified ``\textbf{Grounding} - \textbf{Optimization} - \textbf{Refinement}'' pipeline. 
(1) \textit{\textbf{Verifiable Reasoning Environment Construction}} exposes the internal structure of logical reasoning and makes each reasoning path formally checkable. It decomposes a problem into complementary reasoning views with explicit intermediate judgments and proof obligations, and grounds them into Lean4 for verification. 
(2) \textit{\textbf{Logical Structured Reasoning Optimization}} optimizes different reasoning states at the granularity they require. It combines global supervision over the complete response with path-specific local feedback. The global reward evaluates answer correctness and output structure, while local rewards capture path correctness, proof validity, and consistency, dynamically providing finer-grained optimization signals to the corresponding reasoning paths.
(3) \textit{\textbf{Verifier-Guided Iterative Refinement}} turns verification failures into actionable signals for further reasoning. It feeds proof states, remaining goals, and error messages back to the model, enabling failed reasoning to be diagnosed, revised, and re-verified through an iterative loop.
We evaluate VeriLogic on six logical reasoning benchmarks across three backbone models. VeriLogic consistently improves over the strongest corresponding baselines, with average gains of 1.83\%, 6.39\%, and 5.00\% on Qwen3-8B, Qwen3-4B, and Llama-3.1-8B, respectively. Extensive ablations further confirm the the effectiveness of path-specific advantage routing and verifier-guided refinement.
Overall, these results suggest that the key to improving logical reasoning is not only to reward correct answers, but to make the reasoning process itself explicit, verifiable, and optimizable.

The contributions of this work are summarized as follows:

\begin{itemize}[leftmargin=*,itemsep=0pt,topsep=0pt]

\item \textbf{A new RL paradigm for logical reasoning}. We deeply integrate reinforcement learning with logical reasoning by making the reasoning process both \textbf{traceable} and \textbf{verifiable}. Instead of optimizing only the final answer, VeriLogic exposes multi-path reasoning traces and directly optimizes their intermediate states with verifier-grounded feedback, forming a new RL framework for process-aware logical reasoning.

\item \textbf{A unified ``Grounding - Optimization - Refinement'' RL framework}. To address the challenges of exposing reasoning structure, optimizing heterogeneous reasoning states, and recovering from reasoning failures, VeriLogic introduces three coordinated modules: \textit{\textbf{Verifiable Reasoning Environment Construction}} to ground reasoning into explicit and checkable traces, \textit{\textbf{Logical Structured Reasoning Optimization}} to combine global and local rewards for state-dependent optimization, and \textit{\textbf{Verifier-Guided Iterative Refinement}} to diagnose and repair failed reasoning with verifier feedback.

\end{itemize}
\section{Background and Related Work}
\label{sec:related-work}

\paragraph{LLM logical reasoning.}
Existing work improves LLM reasoning mainly through structured prompting, test-time scaling, and symbolic augmentation (see also modular adapter routing~\citep{li2026lora} for parameter-efficient adaptation). Prompting methods such as Chain-of-Thought, Self-Consistency, Tree-of-Thoughts, Least-to-Most, Plan-and-Solve, and Graph-of-Thoughts encourage models to generate intermediate reasoning steps, aggregate multiple paths, plan ahead, or decompose complex problems before answering \citep{wei2022chain,wang2022self,yao2023tree,zhou2022least,wang2023plan,besta2024graph}. Recent reasoning models further scale this idea with longer reasoning chains, test-time compute, and reinforcement learning from outcome correctness \citep{jaech2024openai,guo2025deepseek,team2025kimi,he2025skywork,guan2025rstar}. For logical reasoning, prior work introduces more explicit structures, including cumulative intermediate conclusions, determinacy-oriented reasoning, logic-augmented contexts, and semiotic-square-guided multi-perspective analysis \citep{zhang2023cumulative,sun2024determlr,liu2025logic,zhang2026semantic}. Another line translates natural-language premises into formal or symbolic representations and invokes external solvers, programs, or provers, such as Logic-LM, LINC, SymbCoT, PAL, CLOVER, and Aristotle \citep{pan2023logic,olausson2023linc,xu2024faithful,gao2023pal,ryu2025divide,xu2025aristotle,yang2026madlogic,mo2026openworld}. While these methods improve logical reasoning through prompting or inference-time tools, they usually do not train the model policy with formally verified feedback from its own generated reasoning.

\paragraph{Verifier-guided reasoning.}
Process-supervision methods evaluate intermediate reasoning steps rather than only final answers, using learned critics, human step labels, or automatically constructed process supervision \citep{uesato2022solving,lightman2024let,wang2024math,luo2024improve,zheng2025survey,liu2026dapo}. However, such signals remain probabilistic and may overlook subtle logical errors. Symbolic verifiers provide stricter feedback by checking reasoning against formal semantics~\citep{jia2025verification}. LogicReward uses theorem-prover-based signals for step-wise logical supervision in natural-language reasoning \citep{xu2026logicreward}, while recent Lean-based systems exploit proof-assistant feedback for formal theorem proving through informal-to-formal guidance, expert iteration, synthetic proof data, compiler feedback, subgoal decomposition, self-correction, copilot-style proof assistance, and reinforcement learning \citep{jiang2022draft,xin2024deepseek,lin2025lean,ying2024internlm,song2024lean,ren2025deepseek,wang2025kimina,lin2025goedel,lin2026goedel,ji2025leanabell,chen2025seed}. Different from these works, our setting focuses on truth-value prediction over natural-language premise sets, where the model must distinguish \texttt{True}, \texttt{False}, and \texttt{Unknown} rather than merely produce a complete formal proof. VeriLogic therefore uses Lean4 feedback to verify the Positive and Negative Paths, while the Unknown Path receives no proof-verification reward. The resulting local credits are restricted to their aligned structured reasoning regions rather than broadcast uniformly across a response.

Beyond formal deduction, social-intelligence research spans affective stance understanding~\citep{zheng2026surfacing}, popularity prediction~\citep{wang2026seeingwiderjointspatiotemporal}, and social simulation~\citep{zhang2026coupling,zhang2026intervensiminterventionawaresocialnetwork,song2026socialintelligence}. AI preferences may also diverge from real user engagement~\citep{zhang2026too}. Structured reasoning is also explored in multi-object tracking~\citep{song2025temporal,song2026hypergraph}.

\section{Methodology}
\label{sec:method}

\begin{figure*}[t!]
\centering
\includegraphics[width=0.98\textwidth]{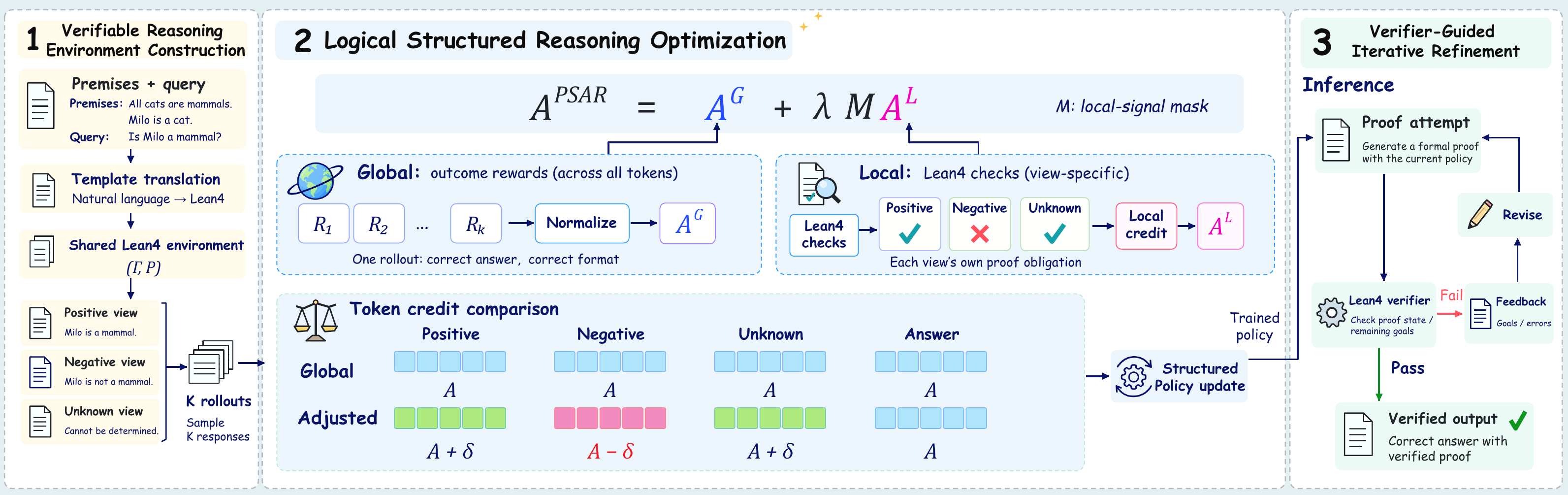}
\caption{\textbf{Overview of VeriLogic.}
VeriLogic follows a unified Grounding--Optimization--Refinement pipeline: it constructs explicit and verifiable reasoning paths, combines global outcome supervision with path-specific local advantages through PSAR-GRPO, and iteratively repairs unsuccessful proofs using Lean4 feedback at inference time.}
\label{fig:pipeline}
\end{figure*}

VeriLogic turns internal logical reasoning into explicit and verifiable learning objects and organizes the overall process into a unified ``\textbf{Grounding}--\textbf{Optimization}--\textbf{Refinement}'' framework. \textit{\textbf{Verifiable Reasoning Environment Construction}} exposes distinct reasoning paths and their proof targets, establishing a formally checkable reasoning space. \textit{\textbf{Logical Structured Reasoning Optimization}} employs \textbf{Path-Specific Advantage Routing GRPO (PSAR-GRPO)} to combine the global outcome advantage with local advantages associated with different reasoning traces, allowing distinct correctness states within the reasoning process to receive corresponding fine-grained optimization signals. \textit{\textbf{Verifier-Guided Iterative Refinement}} leverages proof states and error messages to diagnose and repair failed reasoning. Figure~\ref{fig:pipeline} illustrates the overall VeriLogic pipeline, while Algorithm~\ref{alg:overall-verilogic} summarizes the complete procedure.

\subsection{Verifiable Reasoning Environment Construction}
\label{sec:environment}

\paragraph{Explicit reasoning paths.}
We first formulate logical reasoning as a three-way task with explicit reasoning paths. Each input example is represented as a triple $(C,q,y)$, where $C$ is a natural-language context consisting of facts and rules, $q$ is a queried proposition, and $y\in{\texttt{True},\texttt{False},\texttt{Unknown}}$ denotes the target label. Rather than directly predicting only the final label, we explicitly construct three reasoning paths corresponding to the three possible logical states---Positive, Negative, and Unknown---allowing the model to reason over each possibility separately.

\paragraph{Formal grounding and proof targets.}
To make the provable reasoning paths formally verifiable, we define a mapping $\tau$ that translates the natural-language problem into a Lean4 premise set and a target proposition:
\begin{equation}
\tau(C,q)=(\Gamma,P).
\label{eq}
\end{equation}
Here, $\Gamma$ denotes the Lean4 premise set constructed from the context $C$, and $P$ is the formal proposition semantically aligned with the query $q$. Based on this grounding, the Positive and Negative Paths are associated with the proof targets $P$ and $\neg P$, respectively, and generate formal proofs aligned with their corresponding reasoning directions. The Unknown Path reasons about why the available information is insufficient to support either provable conclusion and therefore does not require a formal proof target.

This construction decomposes an otherwise implicit logical judgment into multiple reasoning paths with explicit semantics and proof targets, allowing different paths to be verified independently and providing the basis for subsequent fine-grained policy optimization over different reasoning states. The complete definition is provided in Appendix~\ref{app:three-path-semantics}.

\subsection{Logical Structured Reasoning Optimization}
\label{sec:optimization}

\paragraph{Structured on-policy verification.}
Given the explicit and verifiable reasoning paths constructed above, we further use verifier feedback to optimize different reasoning states through structured policy learning. Specifically, we train the policy with Group Relative Policy Optimization (GRPO) \citep{shao2024deepseekmath} and integrate Lean4 verification directly into the on-policy rollout loop. For each problem, the current policy samples a group of responses, each containing multiple reasoning paths and their corresponding proofs, with different reasoning paths represented as separately addressable semantic regions. Lean4 checks the generated proofs, allowing reasoning paths and proofs with different correctness states within the same response to be evaluated separately, and the resulting verifier feedback is then used to update the current policy. The complete target mapping and output constraints are provided in Appendix~\ref{app:A}.

\paragraph{Global outcome advantage.}
To provide task-level supervision over the complete reasoning outcome, we first construct a global reward at the response level. For each input, the model explicitly generates three reasoning paths: Positive, Negative, and Unknown. The Positive Path consists of reasoning that supports proposition $P$ together with its corresponding proof, while the Negative Path consists of reasoning that supports $\neg P$ and its corresponding proof. The Unknown Path reasons about why the available information is insufficient to support either provable conclusion and does not contain a formal proof. After completing the three paths, the model synthesizes their reasoning outcomes and autonomously produces the final three-way answer.

The global reward considers final-answer correctness and structured-output validity. For a prompt $u$ with rollout group $\mathcal{G}_u$, we further normalize the response-level rewards within the group to obtain the global advantage:
\begin{equation}
R_i^{\mathrm{G}}
=
\lambda_a R_{\mathrm{ans}}(o_i,y)
+
\lambda_f R_{\mathrm{fmt}}(o_i),
\qquad
A_i^{\mathrm{G}}
=
\frac{R_i^{\mathrm{G}}-\mu_{u,G}}
{\sigma_{u,G}+\epsilon}.
\label{eq:global-reward-advantage}
\end{equation}
Here, $R_{\mathrm{ans}}$ measures whether the final three-way answer matches the target label, while $R_{\mathrm{fmt}}$ measures whether the complete response satisfies the predefined structured-output constraints. $\mu_{u,G}$ and $\sigma_{u,G}$ denote the mean and standard deviation of the global rewards within the rollout group, respectively. The resulting global advantage is broadcast to all valid tokens in the response, providing a unified task-level optimization signal for the complete reasoning process. Since this shared signal captures only the overall outcome and cannot distinguish correctness differences among the three reasoning paths, we further introduce path-level local advantages for fine-grained optimization.

\paragraph{Path-specific local advantage.}
While the global advantage supervises the overall outcome of a response, it cannot distinguish correctness differences among the reasoning paths within the same response. We therefore introduce \textbf{Path-Specific Advantage Routing GRPO (PSAR-GRPO)}, which augments the global advantage with an independently constructed local advantage for each reasoning path. Let
\[
\mathcal{R}=\{\mathrm{pos},\mathrm{neg},\mathrm{unk}\}
\]
denote the Positive, Negative, and Unknown Paths, respectively. For path $r$ in response $i$, we define the path-level local reward as
\begin{equation}
R_{i,r}^{\mathrm{L}}
=
\lambda_p R_{\mathrm{path}}(o_i^{(r)},y)
+
\mathbb{I}[r\in\{\mathrm{pos},\mathrm{neg}\}]
\lambda_v R_{\mathrm{ver}}(o_i^{(r)})
+
\lambda_c R_{\mathrm{cons}}(o_i^{(r)},o_i),
\label{eq:local-reward}
\end{equation}
where $R_{\mathrm{path}}$ measures the correctness of the reasoning state expressed by the current path, $R_{\mathrm{ver}}$ evaluates the validity of the corresponding proof in the Positive and Negative Paths according to Lean4 verification, and $R_{\mathrm{cons}}$ measures consistency among the path reasoning, proof outcome, and final decision. Since the Unknown Path contains no formal proof, its local reward does not include the verification term.

For a prompt $u$ with rollout group $\mathcal{G}_u$, we further normalize the local rewards independently within each path type to obtain path-specific relative advantages:
\begin{equation}
A_{i,r}^{\mathrm{L}}
=
\begin{cases}
\dfrac{R_{i,r}^{\mathrm{L}}-\mu_{u,r}}
{\sigma_{u,r}+\epsilon},
& \sigma_{u,r}>0,\\
0,
& \sigma_{u,r}=0,
\end{cases}
\label{eq:local-advantage}
\end{equation}
where $\mu_{u,r}$ and $\sigma_{u,r}$ denote the mean and standard deviation of the local rewards for path $r$ within the rollout group, respectively. Through this path-wise relative normalization, the Positive, Negative, and Unknown Paths can receive independent local optimization signals according to their own reasoning correctness, proof validity, and consistency, without being directly determined by the correctness states of other paths in the same response. We then combine these path-specific local advantages with the global advantage and route them to their corresponding reasoning regions for policy optimization.

\paragraph{Policy objective.}
Finally, we combine the global advantage with the path-specific local advantages to construct the token-level advantage used for policy optimization. Let $M_{i,t,r}=1$ if and only if token $t$ belongs to the content span of path $r$ in response $i$. We define
\begin{equation}
A_{i,t}^{\mathrm{PSAR}}
=
A_i^{\mathrm{G}}
+
\lambda
\sum_{r\in\mathcal{R}}
M_{i,t,r}A_{i,r}^{\mathrm{L}},
\label{eq:psar-advantage}
\end{equation}
where $A_i^{\mathrm{G}}$ is broadcast over the complete response, whereas the path-specific advantage $A_{i,r}^{\mathrm{L}}$ acts only on its corresponding reasoning region through the path mask. The complete response therefore remains supervised by the global task outcome, while individual reasoning paths receive distinct local corrections according to their own correctness states.

We then use the resulting token-level advantages in the clipped GRPO objective with KL regularization:
\begin{equation}
\begin{aligned}
\mathcal{L}_{\mathrm{PSAR\text{-}GRPO}}(\theta)
=&-\frac{1}{K}\sum_{i=1}^{K}\frac{1}{|\mathcal{T}_i|}
\sum_{t\in\mathcal{T}_i}
\min\Big(
\rho_{i,t}(\theta)A_{i,t}^{\mathrm{PSAR}},\\
&\mathrm{clip}\!\left(
\rho_{i,t}(\theta),
1-\epsilon_{\mathrm{clip}},
1+\epsilon_{\mathrm{clip}}
\right)
A_{i,t}^{\mathrm{PSAR}}
\Big)
+\beta D_{\mathrm{KL}}\!\left(
\pi_{\theta}\,\|\,\pi_{\mathrm{ref}}
\right),
\end{aligned}
\label{eq:psar-grpo-objective}
\end{equation}
where
\begin{equation}
\rho_{i,t}(\theta)
=
\frac{
\pi_{\theta}(o_{i,t}\mid x,o_{i,<t})
}{
\pi_{\theta_{\mathrm{old}}}(o_{i,t}\mid x,o_{i,<t})
}.
\end{equation}
Here, $\mathcal{T}_i$ denotes the set of valid tokens in response $i$. Overall, PSAR-GRPO unifies response-level global outcome supervision and path-level local feedback within a single policy objective, while routing fine-grained learning signals to their corresponding reasoning paths.

\subsection{Verifier-Guided Iterative Refinement}
\label{sec:adaptive-tts}

\paragraph{Iterative proof refinement.}
Even after structured policy optimization, the model may still produce invalid or incomplete formal proofs at inference time. To make verifier feedback useful not only for judging proof correctness but also for repairing failed reasoning, we introduce a verifier-guided iterative refinement mechanism at test time. For a proof rejected by Lean4, the model uses the current proof together with the returned error messages and proof state as an updated reasoning context. It then revises the proof accordingly and resubmits the updated result to Lean4 for verification.

\paragraph{Adaptive decision and fallback.}
Let $S_t$ denote the reasoning context at refinement step $t$. For the Positive and Negative Paths that require formal verification, the model selects
\begin{equation}
a_t \sim \pi_\theta(\cdot\mid S_t),
\qquad
a_t\in
\{\textsc{Prove}(P),\textsc{Prove}(\neg P),\textsc{NoProof}\}.
\label{eq:tts-action-space}
\end{equation}
For proof actions, let $G(a_t)=P$ when $a_t=\textsc{Prove}(P)$ and $G(a_t)=\neg P$ when $a_t=\textsc{Prove}(\neg P)$. Lean4 verifies the current proof attempt and returns
\begin{equation}
F_t=
\mathcal{V}(\Gamma,G(a_t),a_t)
\in
\{\textsc{Success},\textsc{Progress},\textsc{Error}\},
\qquad
a_t\neq\textsc{NoProof}.
\label{eq:tts-feedback}
\end{equation}
Here, \textsc{Success} indicates that the proof is accepted, \textsc{Progress} denotes an incomplete proof with a proof state that can support further refinement, and \textsc{Error} indicates a formal error. For an unsuccessful proof attempt, the model updates its reasoning context using $F_t$ and continues refinement, allowing up to three additional repairs after the initial attempt. If the model selects \textsc{NoProof}, or no valid proof is obtained within the given budget, the system falls back to the final three-way prediction generated from the current three reasoning paths.

\subsection{End-to-End Procedure}
\label{sec:overall-algorithm}

Algorithm~\ref{alg:overall-verilogic} summarizes the complete VeriLogic procedure following the unified ``Grounding--Optimization--Refinement'' framework. Natural-language problems are first transformed into verifiable instances with explicit reasoning paths and proof targets. The policy is then optimized over on-policy rollouts using global and path-specific advantages, while verifier feedback is further used at inference time to iteratively refine unsuccessful proofs.

\begin{algorithm}[H]
\caption{Overall Procedure of VeriLogic}
\label{alg:overall-verilogic}
\begin{algorithmic}[1]
\Require Dataset $\mathcal{D}$, policy $\pi_\theta$, verifier $\mathcal{V}$, group size $K$, refinement budget $N_{\max}$
\Ensure Trained policy $\pi_\theta$ and prediction $\hat{y}$

\State $\mathcal{D}_{\mathrm{F}}
\leftarrow
\{(\Gamma,P,y):(\Gamma,P)=\tau(C,q),\ (C,q,y)\in\mathcal{D}\}$
\cmt{Grounding}

\Repeat
    \State Sample $K$ responses with Positive, Negative, and Unknown Paths from $\pi_\theta$
    \State Verify path proofs with $\mathcal{V}$ and compute $A_i^{\mathrm{G}}$, $A_{i,r}^{\mathrm{L}}$, and $A_{i,t}^{\mathrm{PSAR}}$
    \State Update $\pi_\theta$ using Eq.~\eqref{eq:psar-grpo-objective}
    \cmt{Optimization}
\Until{training budget is exhausted}

\State $(\Gamma,P)\leftarrow\tau(C_{\mathrm{test}},q_{\mathrm{test}})$; generate three reasoning paths and initialize $S_1$
\For{$t=1,\ldots,N_{\max}$}
    \State Refine unresolved Positive/Negative proofs using Lean4 feedback and update $S_{t+1}$
    \cmt{Refinement}
\EndFor
\State \Return $\hat{y}$ from the refined reasoning paths
\end{algorithmic}
\end{algorithm}

\section{Experiments}
\label{sec:experiments}

\begin{table*}[t]
\centering
\footnotesize
\setlength{\tabcolsep}{5.0pt}
\renewcommand{\arraystretch}{1.12}
\caption{\textbf{Main results under the Lean4-formalized setting.} Accuracy is reported for each dataset, and Avg. denotes the average over six benchmarks. $\uparrow$ indicates higher is better. $\Delta$ reports VeriLogic minus the strongest corresponding baseline in each column. Averages and differences are computed before rounding.}
\label{tab:main-results}
\resizebox{\textwidth}{!}{%
\begin{tabular}{@{}p{0.12\textwidth}lccccccc@{}}
\toprule
\textbf{Model}
& \textbf{Method}
& \textbf{MultiLogic}
& \textbf{ProofWriter}
& \textbf{FOLIO}
& \textbf{ProverQA}
& \textbf{LogiQA}
& \textbf{e-SNLI}
& \textbf{Avg.} $\uparrow$ \\
\midrule

\multirow{6}{=}{\centering
\qwen\\[-1pt]
\footnotesize Qwen3-8B}
& Base & 92.67 & 85.33 & 88.00 & 89.33 & 91.67 & 93.00 & 90.00 \\
& GRPO & 94.67 & 88.00 & 90.50 & 91.00 & 95.67 & 96.33 & 92.70 \\
& LogicReward & 95.67 & \second{92.67} & \second{92.50} & \second{94.33} & 96.00 & 97.00 & \second{94.70} \\
& LPT & \second{96.00} & 89.33 & 90.50 & 90.67 & \second{96.33} & \second{98.00} & 93.47 \\
& \cellcolor{gray!12}\textbf{VeriLogic (ours)}
& \cellcolor{gray!12}\best{96.67}
& \cellcolor{gray!12}\best{96.67}
& \cellcolor{gray!12}\best{94.50}
& \cellcolor{gray!12}\best{95.33}
& \cellcolor{gray!12}\best{96.67}
& \cellcolor{gray!12}\best{99.33}
& \cellcolor{gray!12}\best{96.53} \\
& $\Delta$ & \posdelta{+0.67} & \posdelta{+4.00} & \posdelta{+2.00} & \posdelta{+1.00} & \posdelta{+0.34} & \posdelta{+1.33} & \posdelta{+1.83} \\
\midrule

\multirow{6}{=}{\centering
\llama\\[-1pt]
\footnotesize Llama-3.1\\[-2pt]
\footnotesize 8B}
& Base & 55.33 & 45.67 & 57.00 & 48.33 & 58.67 & 74.00 & 56.50 \\
& GRPO & 59.67 & 49.33 & 61.00 & 52.67 & \second{63.33} & 80.33 & 61.06 \\
& LogicReward & 56.33 & \second{52.33} & \second{64.00} & \second{60.67} & 60.00 & \second{81.00} & \second{62.39} \\
& LPT & \second{60.33} & 51.67 & 60.50 & 57.67 & 61.33 & 77.67 & 61.53 \\
& \cellcolor{gray!12}\textbf{VeriLogic (ours)}
& \cellcolor{gray!12}\best{61.00}
& \cellcolor{gray!12}\best{60.67}
& \cellcolor{gray!12}\best{68.00}
& \cellcolor{gray!12}\best{62.67}
& \cellcolor{gray!12}\best{64.67}
& \cellcolor{gray!12}\best{87.33}
& \cellcolor{gray!12}\best{67.39} \\
& $\Delta$ & \posdelta{+0.67} & \posdelta{+8.33} & \posdelta{+4.00} & \posdelta{+2.00} & \posdelta{+1.33} & \posdelta{+6.33} & \posdelta{+5.00} \\
\midrule

\multirow{5}{=}{\centering
\qwen\\[-1pt]
\footnotesize Qwen3-4B}
& Base & 81.67 & 69.67 & 73.50 & 69.33 & 83.67 & 82.33 & 76.70 \\
& GRPO & 85.67 & 74.33 & 77.50 & 74.33 & 88.00 & 86.33 & 81.03 \\
& LPT & \second{88.00} & \second{79.67} & \second{80.00} & \second{78.00} & \second{90.00} & \second{88.67} & \second{84.06} \\
& \cellcolor{gray!12}\textbf{VeriLogic (ours)}
& \cellcolor{gray!12}\best{92.33}
& \cellcolor{gray!12}\best{94.00}
& \cellcolor{gray!12}\best{88.00}
& \cellcolor{gray!12}\best{82.67}
& \cellcolor{gray!12}\best{93.67}
& \cellcolor{gray!12}\best{92.00}
& \cellcolor{gray!12}\best{90.44} \\
& $\Delta$ & \posdelta{+4.33} & \posdelta{+14.33} & \posdelta{+8.00} & \posdelta{+4.67} & \posdelta{+3.67} & \posdelta{+3.33} & \posdelta{+6.39} \\
\bottomrule
\end{tabular}%
}
\end{table*}

\subsection{Experimental Settings}
\paragraph{Dataset.}
We evaluate VeriLogic on six logical reasoning benchmarks under a unified \texttt{True/False/Unknown} setting, covering diverse forms of rule-based and natural-language logical inference. Dataset preprocessing, split details, and statistics are provided in Appendix~\ref{app:B}.

\paragraph{Baselines.}
We compare VeriLogic with Base (the original backbone), GRPO, LogicReward \citep{xu2026logicreward}, and Logical Phase Transitions (LPT) \citep{zhang2026logical}, when available for the corresponding backbone (Appendix~\ref{app:baseline-details}).

\paragraph{Settings.}
We train Qwen3-8B and Qwen3-4B \citep{yang2025qwen3}, and Llama-3.1-8B \citep{grattafiori2024llama} using LoRA fine-tuning \citep{hu2021lora}, and report three-way answer accuracy. At test time, VeriLogic performs up to three additional feedback-guided repairs for unsuccessful proof attempts. Optimization details, decoding parameters, and dataset statistics are provided in Appendix~\ref{app:B}.

\subsection{Main Results}
As shown in Table~\ref{tab:main-results}, VeriLogic achieves the highest average accuracy across all three backbones, outperforming the strongest corresponding baselines by 1.83, 6.39, and 5.00 percentage points on Qwen3-8B, Qwen3-4B, and Llama-3.1-8B, respectively. These results demonstrate consistent gains across different model families and scales. The Non-CoT evaluation in Appendix~\ref{app:additional-analysis} also yields average improvements, although the gains vary across datasets.

\subsection{Ablation Study}
\label{sec:ablation}

\begin{figure}[!t]
\centering
\includegraphics[width=\linewidth]{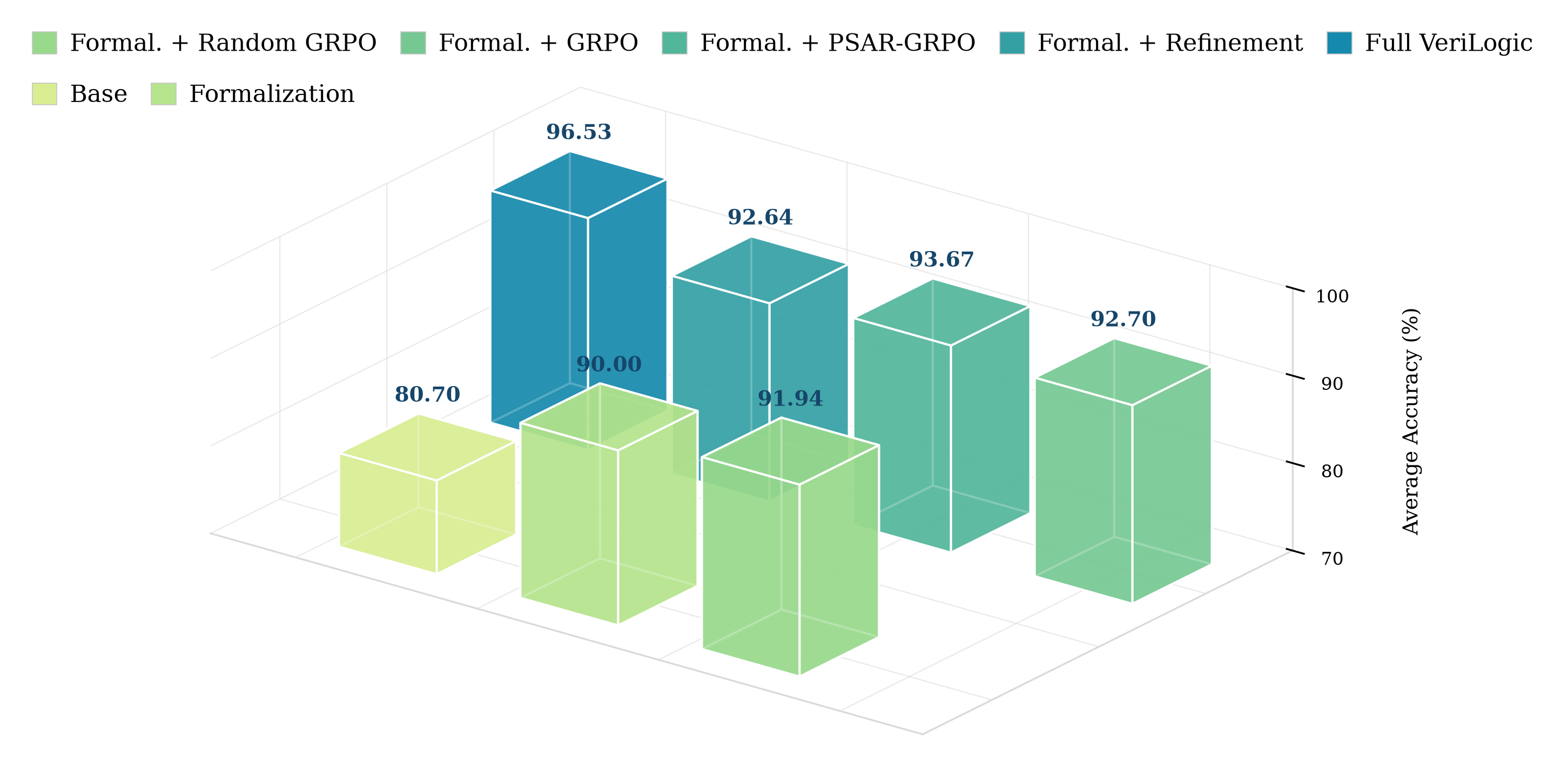}
\caption{\textbf{Component ablation of VeriLogic on Qwen3-8B.} Macro-averaged accuracy over six logical reasoning datasets under formalization, GRPO, PSAR-GRPO, verifier-guided refinement, and full VeriLogic configurations, together with matched controls.}
\label{fig:ablation-3d}
\end{figure}

\paragraph{Ablation setup.}
We isolate the contributions of verifiable formalization, PSAR-GRPO, and verifier-guided iterative refinement on Qwen3-8B. The ablation follows the dependency structure of the framework, with test-time refinement evaluated only on formalized inputs. All settings use the same 1,700-example evaluation set as the main experiments: 300 examples per dataset, except FOLIO with 200. We report macro-averaged accuracy over the six datasets.

\paragraph{Component analysis.}
Figure~\ref{fig:ablation-3d} summarizes the results on Qwen3-8B. \textbf{Formalization} provides the largest single-component gain, increasing average accuracy from 80.70\% to 90.00\%. \textbf{Formalization + GRPO} reaches 92.70\% with the same formalized inputs and global response-level reward, whereas the matched random-reward control reaches 91.94\%. \textbf{PSAR-GRPO} further improves performance to 93.67\%, showing that path-specific local advantages over the Positive, Negative, and Unknown Paths provide finer-grained optimization beyond the global outcome advantage. \textbf{Verifier-guided refinement} reaches 92.64\% with Lean4 feedback. The complete \textbf{VeriLogic} system achieves 96.53\%, a 2.86-point gain over the strongest incomplete configuration, further demonstrating the complementarity of the three components.

\subsection{Analysis}
\label{sec:analysis}

We further analyze complexity robustness, reward validity, path-specific credit assignment, and iterative verifier feedback.

\paragraph{Q1: Complexity--accuracy robustness.}

\begin{wrapfigure}{r}{0.50\columnwidth}
\centering
\includegraphics[width=\linewidth]{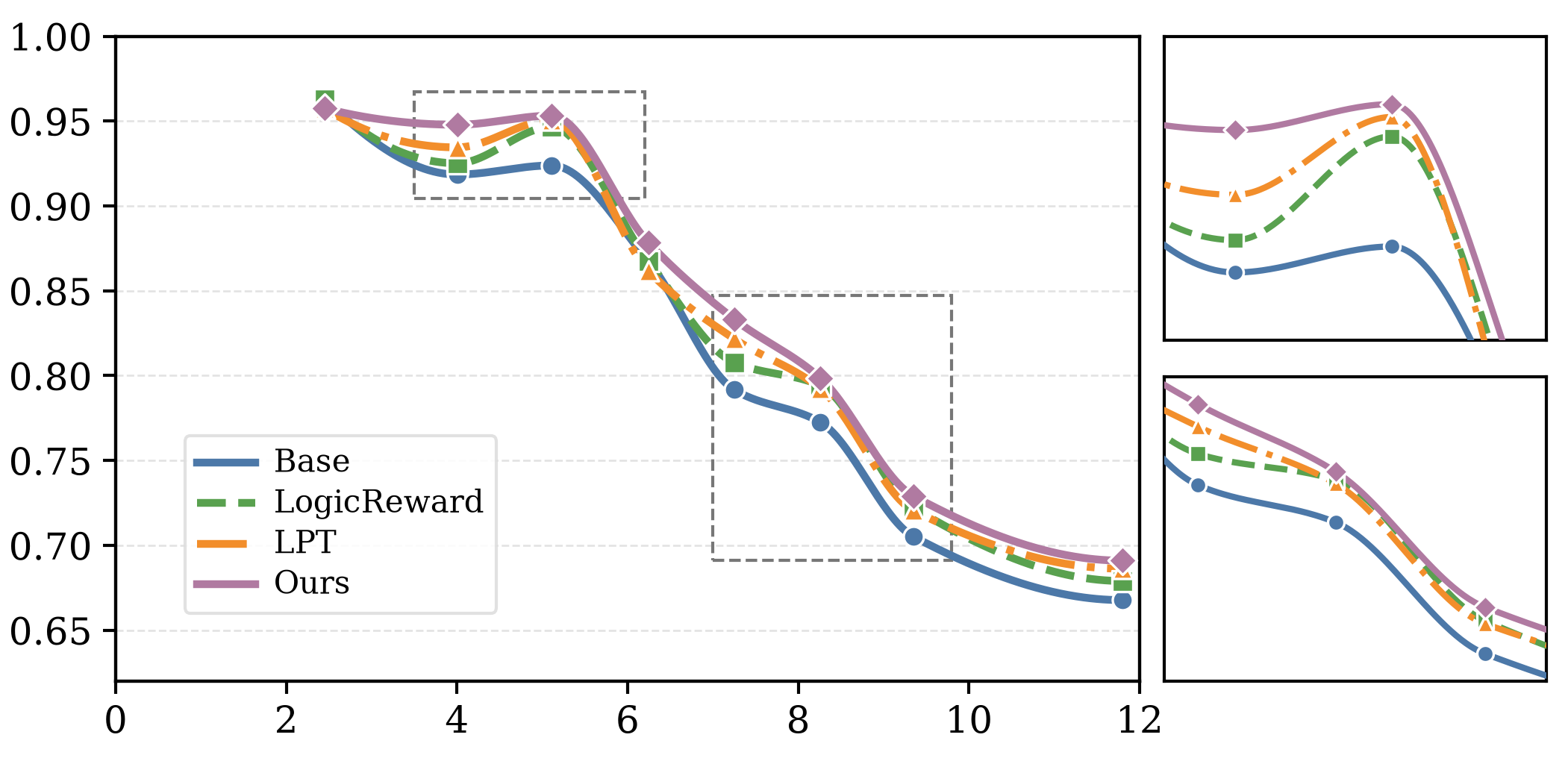}
\captionsetup{justification=centering}
\caption{\textbf{LoCM complexity curves.}}
\label{fig:lpt-complexity}
\end{wrapfigure}

Following the LoCM definition of Zhang et al.~\citep{zhang2026logical}, we partition test instances into eight complexity quantile bins and report the average accuracy in each bin. Figure~\ref{fig:lpt-complexity} shows how performance changes as problem complexity increases. We conduct this analysis on the test set provided by LoCM, rather than the test set used in our main experiments.

All methods exhibit noticeable accuracy drops as LoCM increases, indicating that more complex logical operator compositions and longer reasoning chains remain challenging. VeriLogic achieves the highest or tied-highest accuracy in every complexity bin and maintains comparatively more stable performance in the higher-complexity regions where degradation is more pronounced.

\paragraph{Q2: Task-grounded global reward.}
Under the same formalized inputs, initialization, sampling setup, and optimization budget, \textbf{Formalization + GRPO} uses the task-grounded global response-level reward, while \textbf{Formalization + Random GRPO} replaces it with independently sampled rewards $\tilde{R}_i \sim U[0,1]$. As shown in Figure~\ref{fig:ablation-3d}, the two settings achieve 92.70\% and 91.94\% average accuracy, respectively, a difference of 0.76 percentage points. This comparison shows higher accuracy with task-grounded global rewards in the reported setting, suggesting that the improvement comes not only from policy optimization itself, but also from reward signals that are aligned with the target task. Although the margin is modest, it consistently favors task-grounded supervision under the matched training setup.

\paragraph{Q3: Path-specific advantage routing.}

\begin{wraptable}{r}{0.33\columnwidth}
\centering
\scriptsize
\setlength{\tabcolsep}{3pt}
\renewcommand{\arraystretch}{1.25}
\captionsetup{position=top}
\caption{\textbf{Different routing strategies.}}
\label{tab:routing-ablation}
\begin{tabular}{@{}lrr@{}}
\toprule
\textbf{Routing} & \textbf{Avg. (\%)} & \textbf{$\Delta$} \\
\midrule
\textbf{Base (GRPO)} & \textbf{92.70} & -- \\
Swapped & 72.75 & \textcolor{red!70!black}{$-19.95$} \\
Full-response & 90.76 & \textcolor{red!70!black}{$-1.94$} \\
Random-region & 86.86 & \textcolor{red!70!black}{$-5.84$} \\
\rowcolor{blue!12}\textbf{Correct routing (ours)} & \textbf{93.67} & \textcolor{green!50!black}{\textbf{$+0.97$}} \\
\bottomrule
\end{tabular}
\end{wraptable}

We compare the correct path-specific advantage routing in VeriLogic with several alternative assignments. Figure~\ref{fig:ablation-3d} places the routing result within the overall component ablation, while Table~\ref{tab:routing-ablation} further compares different routing strategies. All settings share the same structured outputs, reward definitions, and training budget. \textbf{Correct routing} applies the local advantages of the Positive, Negative, and Unknown Paths to their corresponding reasoning paths. \textbf{Swapped routing} exchanges the local advantages of the Positive and Negative Paths, \textbf{full-response broadcast} applies each path-level local advantage to the entire response, and \textbf{random-region routing} assigns each local advantage to a random token span with the same length as its intended path.

As shown in Table~\ref{tab:routing-ablation}, correct routing achieves 93.67\% average accuracy, compared with 90.76\% for full-response broadcast, 86.86\% for random-region routing, and 72.75\% for swapped routing. These results indicate that local advantages must not only provide finer-grained supervision, but also be correctly aligned with their corresponding reasoning paths.

\begin{wrapfigure}{r}{0.35\columnwidth}
\centering
\includegraphics[width=0.60\linewidth]{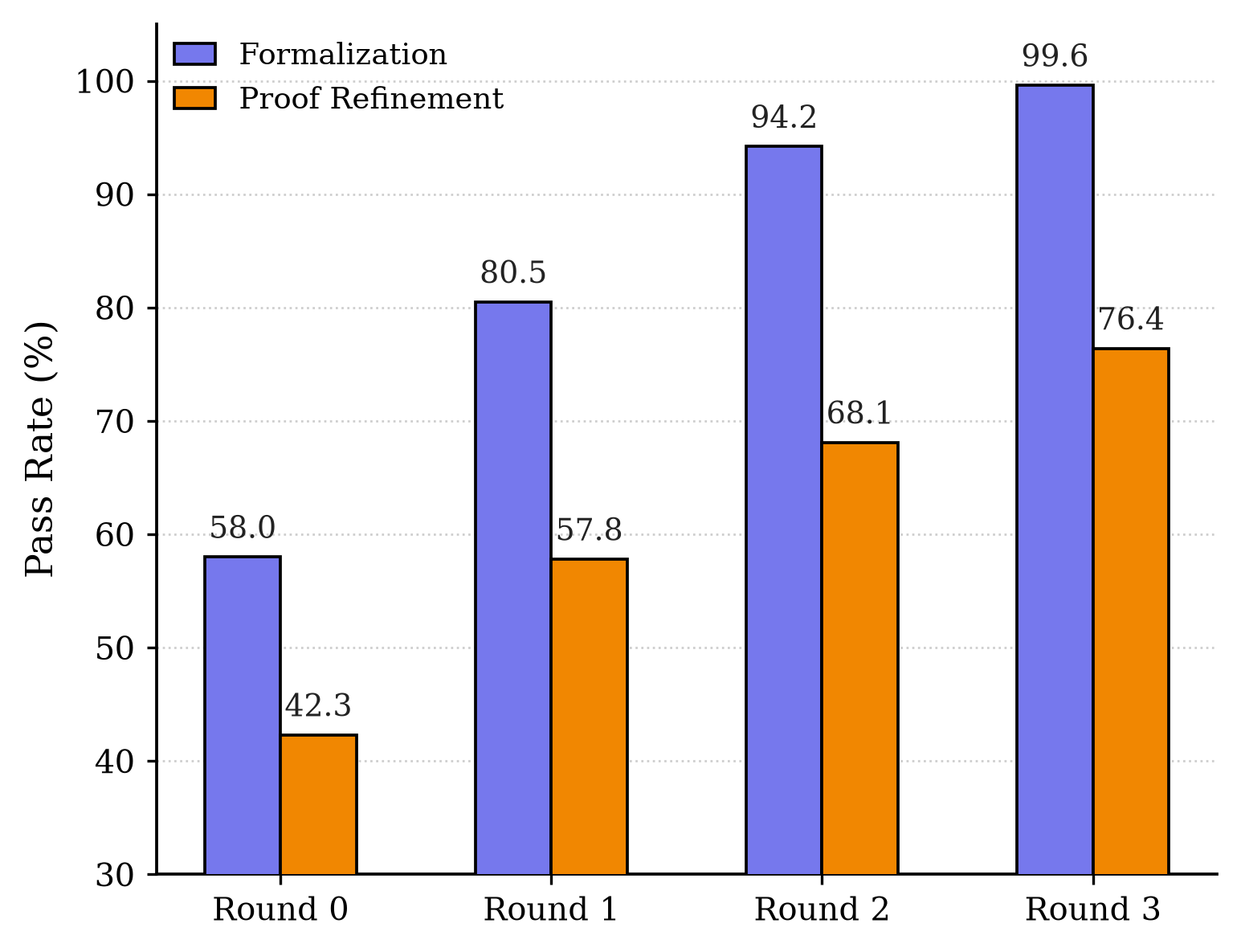}
\captionsetup{hypcap=false}
\caption{\textbf{Verifier feedback by repair round.}}
\label{fig:verifier-feedback}
\vspace{50.53pt}
\vspace{-14pt}
\end{wrapfigure}

\paragraph{Q4: Iterative verifier feedback.}
As shown in Figure~\ref{fig:verifier-feedback}, the formalization pass rate increases from 58.0\% to 99.6\% across verifier-guided repair rounds, while the proof pass rate rises from 42.3\% to 76.4\%. The two curves measure formalization validity and successful proof construction, respectively. Lean4 error messages and proof states are fed back to guide subsequent repairs, yielding consistent gains at both stages. Formalization mainly requires correcting declarations, types, and target statements, and therefore saturates quickly. Proof construction remains more challenging, as the model must further resolve remaining goals and synthesize a complete proof. The gap between the two curves shows that verifier feedback supports both formalization repair and proof refinement while exposing distinct failure modes.

\section{Conclusion}
This study introduces \textbf{VeriLogic}, a reinforcement learning framework that connects natural-language logical reasoning with Lean4 verification. VeriLogic uses \textbf{Verifiable Reasoning Environment Construction} to turn internal reasoning into explicit and verifiable reasoning paths, \textbf{Logical Structured Reasoning Optimization} to combine global supervision over the complete response with path-specific local feedback, and \textbf{Verifier-Guided Iterative Refinement} to iteratively repair unsuccessful proofs using Lean4 proof states and error feedback. Experiments on six logical reasoning benchmarks and three backbone models show consistent improvements in reasoning accuracy, demonstrating the effectiveness of explicit verifiable reasoning, fine-grained path-level optimization, and verifier-guided refinement.

\subsection*{Reproducibility Statement}

All information necessary to reproduce our experiments can be found in Appendices~A--D and the supplementary material.

\subsection*{AI Use Statement}
Generative AI tools were used to convert natural-language logical problems into Lean4 formalizations and to generate candidate reasoning and proof steps for dataset construction. Training and test instances were retained after Lean4 formalization checks, with an additional proof check for \texttt{True} and \texttt{False} instances. AI tools were also used to assist with writing and language polishing; detailed usage is described in the paper. 

\bibliography{main}
\bibliographystyle{ACM-Reference-Format}

\newpage
\appendix
\onecolumn
\raggedbottom

\begin{center}
{\Large \textbf{Appendix}}
\end{center}

\section*{Appendix Overview}
\noindent\textbf{A.} \hyperref[app:A]{Methodological Derivations}\par
\noindent\textbf{B.} \hyperref[app:B]{Experimental Details}\par
\noindent\textbf{C.} \hyperref[app:C]{Additional Analysis}\par
\noindent\textbf{D.} \hyperref[app:D]{Case Study and Prompts}\par

\section{Methodological Derivations}
\label{app:A}
\label{app:method-derivations}
\label{app}

This appendix supplements the Grounding--Optimization--Refinement pipeline of VeriLogic. It clarifies the three reasoning paths, the separation of global and local feedback, and the interpretation of verifier-guided refinement, using the notation of the main text.

\subsection{Formal Grounding and Three-Path Semantics}
\label{app:three-path-semantics}
The mapping $\tau(C,q)=(\Gamma,P)$ in Eq.~\eqref{eq} represents the natural-language premises and query in Lean4. The template specifies entities, predicates, facts, and rules; auxiliary declarations supply the types and definitions required for elaboration. The Positive and Negative Paths share this premise context and use the respective proof targets $P$ and $\neg P$. The Unknown Path reasons about insufficient support for either conclusion and has no formal proof target.

Under the consistent-premise convention, the intended label semantics are
\begin{equation}
 y=\begin{cases}
 \texttt{True},&\Gamma\vdash P,\\
 \texttt{False},&\Gamma\vdash\neg P,\\
 \texttt{Unknown},&\Gamma\nvdash P\ \wedge\ \Gamma\nvdash\neg P.
 \end{cases}
\label{eq:app-three-way-derivability}
\end{equation}
These semantics distinguish a logical state from the outcome of a particular proof attempt. Lean4 acceptance certifies the submitted proposition under $\Gamma$; a rejected or incomplete proof does not establish non-derivability. In particular, two unsuccessful proof attempts do not certify \texttt{Unknown}. Verification also depends on the fidelity of $\tau$: checking a formal proof alone does not establish that the premises faithfully translate the natural-language input.

\paragraph{Structured-output contract.}
A response contains exactly one Positive, Negative, and Unknown Path, in that order, followed by one final answer $\hat y_i\in\{\texttt{True},\texttt{False},\texttt{Unknown}\}$. Each path contains its reasoning and an explicit binary verdict $z_{i,r}$ indicating whether it endorses its associated label, with $y_{\mathrm{pos}}=\texttt{True}$, $y_{\mathrm{neg}}=\texttt{False}$, and $y_{\mathrm{unk}}=\texttt{Unknown}$. Positive and Negative Paths each contain a separately extractable proof field, which may be empty when that path does not claim derivability; a claim of derivability requires a nonempty proof field. The Unknown Path has no proof field. Path boundaries must be unambiguous and nonoverlapping. Missing, duplicated, out-of-order, or unparseable required fields make the format invalid. These structural checks do not require the verdicts to agree with one another or with the final answer: such disagreements are evaluated by the local consistency reward.

\subsection{Global and Path-Level Feedback}
\label{app:reward-details}
The global reward in Eq.~\eqref{eq:global-reward-advantage} evaluates the final answer and the validity of the response structure. Its components are
\begin{equation}
R_{\mathrm{ans}}(o_i,y)=\mathbb I[\hat y_i=y],\qquad
R_{\mathrm{fmt}}(o_i)=\mathbb I[o_i\in\mathcal O_{\mathrm{valid}}],
\label{eq:app-answer-format-reward}
\end{equation}
where $\mathcal O_{\mathrm{valid}}$ is defined by the structured-output contract in Appendix~\ref{app:three-path-semantics}. Format validity concerns the response structure; proof correctness is evaluated separately.

The local reward in Eq.~\eqref{eq:local-reward} distinguishes three sources of feedback. $R_{\mathrm{path}}$ evaluates the logical state expressed by a path against the target label, $R_{\mathrm{ver}}$ evaluates its submitted proof, and $R_{\mathrm{cons}}$ evaluates agreement among the path's reasoning, proof outcome, and final decision. For $r\in\{\mathrm{pos},\mathrm{neg}\}$, write
\begin{equation}
V_{i,r}=\mathbb I[\mathcal V(\Gamma,G_r,\pi_{i,r})=\textsc{Success}],
\qquad G_{\mathrm{pos}}=P,\quad G_{\mathrm{neg}}=\neg P,
\label{eq:app-path-verifier}
\end{equation}
where $\pi_{i,r}$ is the proof submitted in that path. A missing, incomplete, or rejected proof supplies no successful verification evidence. The Unknown Path receives path-correctness and consistency feedback but no verification reward. Thus, correctness of the final answer does not by itself determine the local rewards, and failure to prove either polarity is not treated as a verified Unknown judgment.

\paragraph{Binary scoring specification.}
To make the reward terms explicit, define the gold path indicator $z_r^\star=\mathbb I[y=y_r]$ and the answer-induced indicator $\hat z_{i,r}=\mathbb I[\hat y_i=y_r]$. A binary realization of the local reward components is
\begin{equation}
\begin{aligned}
R_{\mathrm{path}}(o_i^{(r)},y)&=\mathbb I[z_{i,r}=z_r^\star],\\
R_{\mathrm{ver}}(o_i^{(r)})&=V_{i,r},\qquad r\in\{\mathrm{pos},\mathrm{neg}\},\\
R_{\mathrm{cons}}(o_i^{(r)},o_i)&=\mathbb I[z_{i,r}=\hat z_{i,r}]\,C_{i,r},\\
C_{i,r}&=\begin{cases}
\mathbb I[z_{i,r}=V_{i,r}],&r\in\{\mathrm{pos},\mathrm{neg}\},\\
1,&r=\mathrm{unk}.
\end{cases}
\end{aligned}
\label{eq:app-binary-local-rewards}
\end{equation}
Thus, endorsing the gold path or rejecting a non-gold path earns path-correctness credit. A positive derivability claim is proof-consistent only with an accepted proof; an accepted proof contradicts a negative derivability verdict. A negative verdict with no accepted proof is merely non-conflicting, not certified non-derivability: its correctness is assessed separately against the gold label. The Unknown verdict is checked against the final answer for consistency and against the gold label for correctness, without any proof reward. These checks operate on explicit verdicts and proof outcomes; they do not independently verify every natural-language reasoning step.

For example, if $y=\texttt{True}$, $\hat y_i=\texttt{True}$, and $z_{i,\mathrm{pos}}=1$ but its proof fails, the Positive Path receives $R_{\mathrm{path}}=1$, $R_{\mathrm{ver}}=0$, and $R_{\mathrm{cons}}=0$. This separates a correct judgment from an unsupported proof claim. For an unparseable path, its local reward components and token mask are set to zero; an unparseable final answer receives zero answer and consistency rewards. Format violations always receive $R_{\mathrm{fmt}}=0$.

\subsection{Path-Specific Normalization and Routing}
\label{app:psar-grpo}
For a rollout group $\mathcal G_u$ of size $K>1$, the statistics for each path $r\in\{\mathrm{pos},\mathrm{neg},\mathrm{unk}\}$ are
\begin{equation}
\mu_{u,r}=\frac1K\sum_{j\in\mathcal G_u}R_{j,r}^{\mathrm L},\qquad
\sigma_{u,r}=\sqrt{\frac1{K-1}\sum_{j\in\mathcal G_u}
(R_{j,r}^{\mathrm L}-\mu_{u,r})^2}.
\label{eq:app-local-statistics}
\end{equation}
Global statistics are computed in the same way from $R_j^{\mathrm G}$. Normalization compares rewards across responses within the same path type, rather than pooling different paths. As in Eq.~\eqref{eq:local-advantage}, a path with zero reward variance receives zero local advantage.

The mask $M_{i,t,r}$ in Eq.~\eqref{eq:psar-advantage} selects the content tokens of path $r$, including its associated proof when present. The final-answer tokens and other tokens outside the path spans receive only $A_i^{\mathrm G}$. Consequently, local feedback changes the update within its corresponding path while global supervision remains active across the complete response. If $\lambda=0$, or all local advantages vanish, the update reduces to the global-only objective with the same response-level reward.

The resulting token advantages are inserted directly into the clipped, KL-regularized objective in Eq.~\eqref{eq:psar-grpo-objective}. Rewards, normalization statistics, and masks are held fixed during each policy update. This routing controls where local credit is applied; it does not imply independent model parameters or independent reasoning paths.

\subsection{Verifier-Guided Refinement and Final Prediction}
\label{app:refinement-details}
Following Section~\ref{sec:adaptive-tts}, inference allows one initial proof attempt and up to three additional repairs. For an unsuccessful Positive or Negative proof, the next context includes the current attempt together with Lean4's error messages, remaining goals, and available hypotheses:
\begin{equation}
S_{t+1}=\mathrm{Concat}(S_t,a_t,F_t).
\label{eq:app-tts-context-update}
\end{equation}
The statuses \textsc{Success}, \textsc{Progress}, and \textsc{Error} retain the meanings given in Eq.~\eqref{eq:tts-feedback}. Only \textsc{Success} supplies an accepted proof; the other statuses provide information for revision.

An accepted proof supports the corresponding Positive or Negative conclusion. If the model selects \textsc{NoProof}, or the repair budget is exhausted without a valid proof, the final three-way prediction is generated from the current reasoning paths, as in Algorithm~\ref{alg:overall-verilogic}. Neither condition automatically implies \texttt{Unknown}. Final-answer accuracy and successful formal verification therefore measure distinct aspects of the output.

\section{Experimental Details}
\label{app:B}
\label{app:experimental-settings}

This section summarizes the experimental settings used in our main experiments. It covers dataset preprocessing and splits, dataset statistics, optimization and generation settings, and baseline sources.

\subsection{Datasets, Preprocessing, and Splits}
\label{app:data-details}

We evaluate on six logical reasoning datasets: MultiLogic, which covers diverse logical forms and rule combinations \citep{patel2024multi}; ProofWriter, which evaluates rule-based natural language reasoning with facts, rules, and proof structures \citep{tafjord2021proofwriter}; FOLIO, which provides natural language reasoning examples with first-order logic annotations \citep{han2024folio}; ProverQA, a large-scale first-order logic reasoning dataset constructed with the assistance of symbolic provers \citep{qi2025large}; LogiQA, a reading-comprehension-style logical reasoning benchmark derived from examination scenarios \citep{liu2020logiqa}; and e-SNLI, a natural-language reasoning dataset with entailment, contradiction, and neutral labels \citep{camburu2018snli}. For FOLIO, we additionally refer to the verified annotation resources released by \citet{brunello2026fixing}, which address annotation issues in the original dataset. These datasets are also connected to a broader line of logical-reasoning benchmarks and analyses that test proof planning, rule application, multi-step proof construction, negation, and systematic logical patterns \citep{saparov2022language,zhao2023explicit,zheng2024exploring,parmar2024logicbench,jin2023cladder}.

\paragraph{Label normalization.}
All datasets are normalized into a unified three-way reasoning format:
\[
\text{premises}+\text{question}
\rightarrow
\{\textsc{True},\textsc{False},\textsc{Unknown}\}.
\]
For datasets whose original label space does not explicitly include 
\textsc{Unknown}, we use a stronger model to assist label normalization. 
This step separates refutable \textsc{False} cases from underdetermined 
\textsc{Unknown} cases, reducing the risk of mixing unverifiable or 
underspecified examples into the \textsc{False} class.

\paragraph{Splits and retention.}
All six datasets are used for both training and evaluation, with randomly sampled training and evaluation subsets constructed from each dataset. Dataset statistics after preprocessing are shown in Table~\ref{tab:dataset-training-details}.

Training and evaluation use the same retention criteria: all instances must pass Lean4 formalization checks; \texttt{True} and \texttt{False} instances additionally require an accepted proof of $P$ and $\neg P$, respectively, while \texttt{Unknown} instances are retained after formalization.

\subsection{Training Configuration}
\label{app:training-details}

\paragraph{Optimization.}
The full VeriLogic model is optimized with PSAR-GRPO using LoRA fine-tuning on 2 H100 GPUs. Table~\ref{tab:dataset-training-details} summarizes the dataset statistics and main training hyperparameters.

\begin{table}[!htbp]
\centering
\small
\setlength{\tabcolsep}{5pt}
\renewcommand{\arraystretch}{1.12}
\caption{Dataset statistics and training details of VeriLogic.}
\label{tab:dataset-training-details}

\textbf{(a) Dataset sizes after preprocessing.}

\vspace{3pt}
\resizebox{\linewidth}{!}{%
\begin{tabular}{lrrrrrrr}
\toprule
\textbf{Dataset}
& \textbf{e-SNLI}
& \textbf{FOLIO}
& \textbf{LogiQA}
& \textbf{MultiLogic}
& \textbf{ProofWriter}
& \textbf{ProverQA}
& \textbf{Total} \\
\midrule
\textbf{Training examples}
& 458 & 351 & 527 & 380 & 1,727 & 2,533 & 5,976 \\
\textbf{Test examples}
& 300 & 200 & 300 & 300 & 300 & 300 & 1,700 \\
\bottomrule
\end{tabular}%
}

\vspace{8pt}

\textbf{(b) Training details.}

\vspace{3pt}
\begin{tabular}{lp{0.70\textwidth}}
\toprule
\textbf{Item} & \textbf{Setting} \\
\midrule
Backbone models
& Qwen3-8B and Qwen3-4B; Llama-3.1-8B \\

Training algorithm
& PSAR-GRPO with LoRA fine-tuning \\

LoRA configuration
& Rank 16 / alpha 32 for all three backbones \\

LoRA target modules
& All linear modules \\

Precision
& bfloat16 \\

Group size
& 8 sampled completions per prompt \\

Effective batch size
& 32 \\

Optimization
& 3 epochs, learning rate $1\times10^{-5}$, $\beta=0.04$ \\

Generation
& Temperature 0.7, maximum sequence length 8192, maximum completion length 4096 \\

\bottomrule
\end{tabular}

\end{table}

\subsection{Baseline Details}
\label{app:baseline-details}

Base denotes the original backbone model without fine-tuning. LogicReward uses publicly released weights when available. LPT is reproduced under our experimental setting. Missing baseline entries are omitted when no comparable checkpoint or setting is available for the corresponding backbone.

\section{Additional Analysis}
\label{app:C}
\label{app:additional-analysis}
This section evaluates whether the improvements from Lean4-guided training remain visible when explicit reasoning traces are removed from the model output. In the Non-CoT setting, the model directly generates the final three-way answer without generating intermediate reasoning traces.

\begin{table}[!htbp]
\centering
\footnotesize
\setlength{\tabcolsep}{5.0pt}
\renewcommand{\arraystretch}{1.12}
\caption{\textbf{Main results under the Lean4-formalized Non-CoT setting.} Accuracy is reported for each dataset, and Avg. is the average over the six reported benchmarks. $\uparrow$ indicates that higher values are better. $\Delta$ reports VeriLogic minus the strongest baseline for each column.}
\label{tab:non-cot-main-results}
\resizebox{\textwidth}{!}{%
\begin{tabular}{@{}p{0.12\textwidth}lccccccc@{}}
\toprule
\textbf{Model}
& \textbf{Method}
& \textbf{MultiLogic}
& \textbf{ProofWriter}
& \textbf{FOLIO}
& \textbf{ProverQA}
& \textbf{LogiQA}
& \textbf{e-SNLI}
& \textbf{Avg.} $\uparrow$ \\
\midrule

\multirow{6}{=}{\centering
\qwen\\[-1pt]
\footnotesize Qwen3-8B}
& Base & 65.33 & 64.67 & 63.50 & 59.67 & 72.00 & 89.00 & 69.03 \\
& GRPO & 68.33 & \second{68.00} & \second{67.00} & \second{63.00} & 75.67 & 92.67 & 72.44 \\
& LogicReward & \best{69.67} & 65.67 & \second{67.00} & \second{63.00} & \best{78.00} & \best{93.67} & \second{72.83} \\
& LPT & \second{68.67} & 67.33 & \best{67.50} & 62.67 & \second{76.00} & 92.67 & 72.47 \\
& \cellcolor{gray!12}\textbf{VeriLogic (ours)}
& \cellcolor{gray!12}\second{68.67}
& \cellcolor{gray!12}\best{73.67}
& \cellcolor{gray!12}\second{67.00}
& \cellcolor{gray!12}\best{63.33}
& \cellcolor{gray!12}\second{76.00}
& \cellcolor{gray!12}\second{93.00}
& \cellcolor{gray!12}\best{73.61} \\
& $\Delta$ & \negdelta{-1.00} & \posdelta{+5.67} & \negdelta{-0.50} & \posdelta{+0.33} & \negdelta{-2.00} & \negdelta{-0.67} & \posdelta{+0.78} \\

\midrule

\multirow{6}{=}{\centering
\llama\\[-1pt]
\footnotesize Llama-3.1\\[-2pt]
\footnotesize 8B}
& Base & 29.67 & 46.33 & 39.50 & 40.67 & 33.00 & 51.67 & 40.14 \\
& GRPO & 32.33 & 50.67 & 43.00 & 44.67 & 36.33 & 55.00 & 43.67 \\
& LogicReward & 39.67 & \second{57.33} & \best{50.50} & \best{48.33} & \second{37.00} & \second{66.67} & \second{49.92} \\
& LPT & \second{40.33} & 54.00 & 44.50 & 45.33 & 33.00 & 63.67 & 46.81 \\
& \cellcolor{gray!12}\textbf{VeriLogic (ours)}
& \cellcolor{gray!12}\best{40.67}
& \cellcolor{gray!12}\best{58.00}
& \cellcolor{gray!12}\second{49.50}
& \cellcolor{gray!12}\second{46.67}
& \cellcolor{gray!12}\best{38.67}
& \cellcolor{gray!12}\best{68.00}
& \cellcolor{gray!12}\best{50.25} \\
& $\Delta$ & \posdelta{+0.33} & \posdelta{+0.67} & \negdelta{-1.00} & \negdelta{-1.67} & \posdelta{+1.67} & \posdelta{+1.33} & \posdelta{+0.33} \\

\midrule

\multirow{5}{=}{\centering
\qwen\\[-1pt]
\footnotesize Qwen3-4B}
& Base & 62.33 & 59.67 & 60.50 & 53.33 & 72.33 & 84.67 & 65.47 \\
& GRPO & \second{66.67} & \second{63.67} & 64.50 & 57.33 & \best{77.00} & \best{88.00} & 69.53 \\
& LPT & \best{67.33} & \second{63.67} & \second{65.50} & \second{58.67} & \second{75.33} & \second{87.67} & \second{69.69} \\
& \cellcolor{gray!12}\textbf{VeriLogic (ours)}
& \cellcolor{gray!12}66.00
& \cellcolor{gray!12}\best{67.00}
& \cellcolor{gray!12}\best{67.00}
& \cellcolor{gray!12}\best{59.67}
& \cellcolor{gray!12}\second{75.33}
& \cellcolor{gray!12}\second{87.67}
& \cellcolor{gray!12}\best{70.44} \\
& $\Delta$ & \negdelta{-1.33} & \posdelta{+3.33} & \posdelta{+1.50} & \posdelta{+1.00} & \negdelta{-1.67} & \negdelta{-0.33} & \posdelta{+0.75} \\

\bottomrule
\end{tabular}%
}
\end{table}

As shown in Table~\ref{tab:non-cot-main-results}, VeriLogic achieves the best average accuracy for all three backbones in the Non-CoT setting, improving over the strongest corresponding baselines by 0.78, 0.33, and 0.75 points for Qwen3-8B, Llama-3.1-8B, and Qwen3-4B, respectively. These results show small average gains under answer-only evaluation, with larger improvements in the CoT setting where the model can express its structured reasoning.

\section{Case Study and Prompts}
\label{app:D}
\label{app:prompts}
This appendix presents a compact formalization case and a simplified prompt template aligned with the Positive, Negative, and Unknown Paths of VeriLogic. The output fields follow Appendix~\ref{app:three-path-semantics}.

\subsection{Compact End-to-End Lean4 Verification Case}
\label{app:formalization-example}

This case study connects natural-language premises, the two Lean4 proof targets, and the three-path judgment leading to a final answer.

\tcbset{
    colback=blue!5,
    colframe=blue!75,
    fonttitle=\bfseries,
    width=\textwidth,
    boxrule=0.5mm,
    arc=3mm,
    breakable,
    enhanced,
    before skip=8pt,
    after skip=8pt
}

\begin{tcolorbox}[title=Compact End-to-End Lean4 Verification Case]\footnotesize

\textbf{Natural-language problem.}
Elian appreciates culture.
Anyone who appreciates culture travels extensively.
Anyone who travels extensively becomes experienced.
If Elian is experienced, then he cannot both lack refinement and be sophisticated.
Elian is sophisticated.
Elian either lacks refinement or has good taste.
Elian cannot have both good taste and bad taste.

\textbf{Question.}
Is the statement ``Elian has bad taste'' true, false, or unknown?

\textbf{Polarity targets.}
\[
P := badTaste(Elian),
\qquad
G^{+}=P,
\qquad
G^{-}=\neg P.
\]

\textbf{Lean4 instance and proof of the negative target.}
\begin{Verbatim}[fontsize=\scriptsize,breaklines=true,breakanywhere=true]
axiom Entity : Type
axiom Elian : Entity

axiom appreciatesCulture : Entity -> Prop
axiom travelsExtensively : Entity -> Prop
axiom becomesExperienced : Entity -> Prop
axiom lacksRefinement : Entity -> Prop
axiom isSophisticated : Entity -> Prop
axiom goodTaste : Entity -> Prop
axiom badTaste : Entity -> Prop

axiom elianAppreciatesCulture :
  appreciatesCulture Elian

axiom cultureToTravel :
  (x : Entity) -> appreciatesCulture x -> travelsExtensively x

axiom travelToExperienced :
  (x : Entity) -> travelsExtensively x -> becomesExperienced x

axiom experiencedExclusive :
  becomesExperienced Elian ->
  ¬ (lacksRefinement Elian ∧ isSophisticated Elian)

axiom elianSophisticated :
  isSophisticated Elian

axiom refinementOrGoodTaste :
  lacksRefinement Elian ∨ goodTaste Elian

axiom tasteExclusive :
  ¬ (goodTaste Elian ∧ badTaste Elian)

theorem elian_not_bad_taste : ¬ badTaste Elian := by
  have hExp : becomesExperienced Elian :=
    travelToExperienced Elian
      (cultureToTravel Elian elianAppreciatesCulture)

  have hNoLack : ¬ lacksRefinement Elian := by
    intro hLack
    exact experiencedExclusive hExp
      ⟨hLack, elianSophisticated⟩

  have hGood : goodTaste Elian := by
    cases refinementOrGoodTaste with
    | inl hLack =>
        exact False.elim (hNoLack hLack)
    | inr hGood =>
        exact hGood

  intro hBad
  exact tasteExclusive ⟨hGood, hBad⟩
\end{Verbatim}

\textbf{Three-path interpretation.}
The Positive Path is \texttt{unsupported}: the premises entail the negation of the query. The Negative Path is \texttt{supported}, with the proof above establishing $\neg P$. The Unknown Path is \texttt{unsupported}, since a definite negative conclusion is available. The empty Positive proof field does not itself establish non-derivability; the negative conclusion is supported by the constructed proof.

\textbf{Final answer.}
The proof derives the negative target:
\[
\Gamma \vdash \neg badTaste(Elian).
\]
Thus the original statement $badTaste(Elian)$ is \texttt{False}:
\[
\boxed{y=\texttt{False}}.
\]

\end{tcolorbox}

\subsection{GRPO Prompt Template}
\label{app:grpo-prompt-template}

The following simplified template specifies the three-path response structure used in the main text; it is not a verbatim training prompt. Each path states whether its associated label is supported. Proofs are nested within the Positive and Negative Paths, while the Unknown Path contains no proof field. The final answer is generated after considering all three paths.

\begin{tcolorbox}[title=Simplified Three-Path GRPO Prompt Template]\footnotesize

\textbf{System instruction.}
\begin{Verbatim}[fontsize=\scriptsize,breaklines=true,breakanywhere=true]
You reason over a Lean4 premise context and a query P.
Assess three paths before producing a final answer:
- Positive: whether the premises support P (True).
- Negative: whether the premises support Not P (False).
- Unknown: whether neither conclusion is supported (Unknown).

For each path, write a short explanation and one verdict:
  supported | unsupported
The verdict endorses or rejects that path's associated label.
A failed proof attempt alone does not establish Unknown.

Proof requirements:
- Place each proof inside its own Positive or Negative Path.
- A supported Positive/Negative verdict requires a nonempty
  proof field targeting P/Not P, respectively.
- An unsupported Positive/Negative verdict may have an empty
  proof field. Never put a proof field in the Unknown Path.
- Use only the supplied premises and complete Lean4 proofs.
  Do not introduce additional axioms or use sorry or admit.
- Do not treat an unsuccessful attempt as verified evidence.

Output exactly one positive, negative, and unknown block,
in that order, followed by one answer. Replace placeholders
and choose one verdict and one final label; do not print the
alternative lists literally. Generate the final label by
considering the three paths and the available proof evidence.

<think>
<positive>
<verdict>supported | unsupported</verdict>
[Brief reasoning about P]
<proof>[Lean4 proof of P, or empty]</proof>
</positive>
<negative>
<verdict>supported | unsupported</verdict>
[Brief reasoning about Not P]
<proof>[Lean4 proof of Not P, or empty]</proof>
</negative>
<unknown>
<verdict>supported | unsupported</verdict>
[Brief reasoning about whether neither conclusion is supported]
</unknown>
</think>
<answer>{"label":"True" | "False" | "Unknown"}</answer>
\end{Verbatim}

\textbf{User prompt.}
\begin{Verbatim}[fontsize=\scriptsize,breaklines=true,breakanywhere=true]
Premise context:
<LEAN4_DECLARATIONS_AND_PREMISES>

Positive theorem statement: <POSITIVE_THEOREM_NAME> : <P>
Negative theorem statement: <NEGATIVE_THEOREM_NAME> : Not (<P>)

Assess the Positive, Negative, and Unknown Paths using this
shared context. Return the required structured response and
one final three-way answer. Complete a theorem with := by
and its proof when claiming that the corresponding goal is
supported.
\end{Verbatim}

\end{tcolorbox}

\end{document}